\documentclass[lettersize,journal]{IEEEtran}
\usepackage{amsmath,amsfonts}
\usepackage{algorithmic}
\usepackage{algorithm}
\usepackage{array}
\usepackage{textcomp}
\usepackage{stfloats}
\usepackage{url}
\usepackage{verbatim}
\usepackage{graphicx}
\usepackage{cite}
\usepackage{multirow}
\usepackage{enumitem}
\usepackage{colortbl}
\usepackage{subcaption}

\definecolor{gray}{gray}{.9}
\usepackage{bm}
\usepackage[colorlinks, linkcolor=red, anchorcolor=red, citecolor=green]{hyperref}

\begin{document}

\title{PRSeg: A Lightweight Patch Rotate MLP Decoder for Semantic Segmentation}

\author{ Yizhe Ma\textsuperscript{\rm 1 $*$}, 
Fangjian Lin\textsuperscript{\rm 1, \rm 2 $*$},
Sitong Wu\textsuperscript{\rm 3},
Shengwei Tian\textsuperscript{\rm 1}, 
Long Yu\textsuperscript{ \rm 1, $\dag$}

\thanks{
\textsuperscript{$*$} Equal contributions
\textsuperscript{$\dag$} Corresponding author. \\
\textsuperscript{\rm 1} School of Software, Xinjiang University, Urumqi, China \\
\textsuperscript{\rm 2} Shanghai AI Laboratory, Shanghai, China\\
\textsuperscript{\rm 3} Baidu Inc.
}
}

\markboth{Journal of \LaTeX\ Class Files,~Vol.~14, No.~8, August~2021}%
{Shell \MakeLowercase{\textit{et al.}}: A Sample Article Using IEEEtran.cls for IEEE Journals}


\maketitle
\begin{abstract}
The lightweight MLP-based decoder has become increasingly promising for semantic segmentation.
However, the channel-wise MLP cannot expand the receptive fields, lacking the context modeling capacity, which is critical to semantic segmentation.
In this paper, we propose a parametric-free patch rotate operation to reorganize the pixels spatially.
It first divides the feature map into multiple groups and then rotates the patches within each group. 
Based on the proposed patch rotate operation, we design a novel segmentation network, named PRSeg, which includes an off-the-shelf backbone and a lightweight Patch Rotate MLP decoder containing multiple Dynamic Patch Rotate Blocks (DPR-Blocks). 
In each DPR-Block, the fully connected layer is performed following a Patch Rotate Module (PRM) to exchange spatial information between pixels. 
Specifically, in PRM, the feature map is first split into the reserved part and rotated part along the channel dimension according to the predicted probability of the Dynamic Channel Selection Module (DCSM), and our proposed patch rotate operation is only performed on the rotated part. 
Extensive experiments on ADE20K, Cityscapes and COCO-Stuff 10K datasets prove the effectiveness of our approach.
We expect that our PRSeg can promote the development of MLP-based decoder in semantic segmentation.
\end{abstract}


\section{Introduction}

\IEEEPARstart{S}{emantic} segmentation aims to predict a semantic label for each pixel in an image. 
With the development of autonomous driving, human-computer interaction, augmented reality, etc., semantic segmentation has received more and more attention.
Over the recent decade, encoder-decoder based segmentation methods have become the dominant models, where the encoder is usually implemented by an off-the-shelf backbone network \cite{ResNet,ViT,swin,Pale}, and the decoder utilizes the spatial and semantic features to generate the pixel-level prediction.

\begin{figure}[tp]
\centering
\includegraphics[width=1\linewidth]{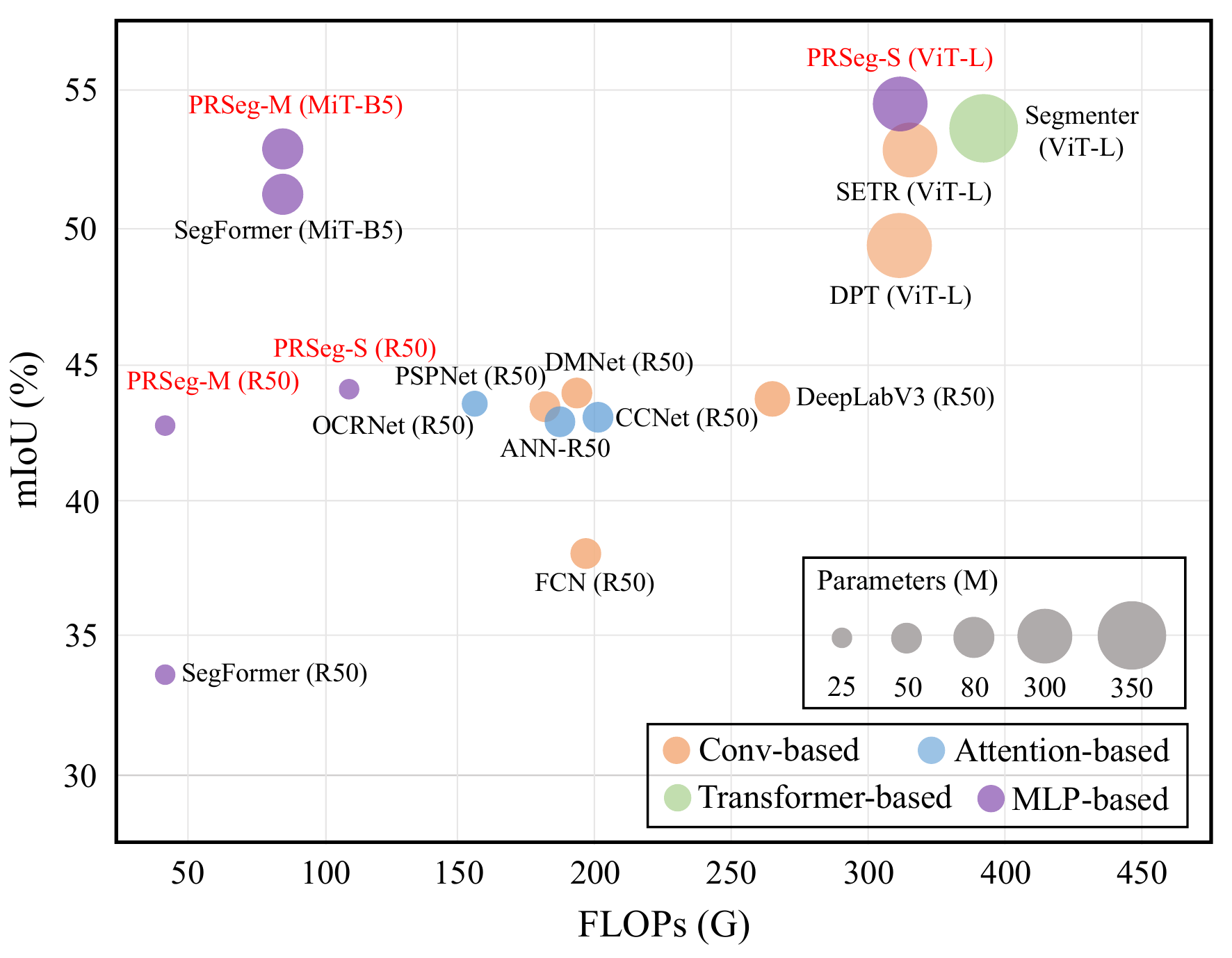} 
\caption{
Model performance vs. model efficiency on the ADE20K validation set. Results are presented for single models with multi-scale inference. Through the MLP decoder, PRSeg achieves a new state-of-the-art 54.16$\%$ mIoU while being more efficient than previous methods.}
\label{compare}
\end{figure}
Many efforts have been made in the design of decoders for semantic segmentation \cite{CGNet, Segmenter, StructToken, FTN}.
According to the basic structure of layers, previous segmentation decoders can be summarized into four categories:
(1) \textit{Convolution-based decoder.} It mainly uses convolution to model contextual relationships, expand the perceptive field, extract local detail features, fuse multi-scale information, etc. For example, PSPNet\cite{pspnet} utilizes Pyramid pooling techniques to model contextual features and expand the perceptive field. Some works\cite{semantic_fpn, SETR, DPT} use top-down connectivity to fuse multi-scale spatial and semantic information to enhance pixel representation and multi-scale object recognition.
(2) \textit{Attention-based decoder.} It means that the decoder uses the attention technique to obtain the similarity between pixels, and provides a focus on the region of interest. OCRNet\cite{OCRNet} uses attention to compute the similarity of pixel features to individual class features, and CCNet\cite{CCNet} uses Criss-Cross Attention to capture long-range dependencies and model rich contextual features. ANN\cite{ann} uses axial attention to capture remote subsurface context, etc.
(3) \textit{Transformer-based decoder}. Some researchers have incorporated the transformer into decoder designs as a result of its ability to model global contextual features, capture long-range dependencies, and provide flexible prototype representations. For example, Segmenter\cite{Segmenter} uses transformer to build global contextual relations and embeds the prototypes of categories into transformer to learn together with feature maps. MaskFormer\cite{maskformer} uses transformer's cross-attention to let the mask token learn the feature map’s category representation. 
(4) \textit{MLP-based decoder.} 
Recently, the MLP-based decoder has received more and more attention, mainly because of two aspects, the lightweight and efficient property of the MLP architecture, and the development of the transformer backbone, which offers global extraction capabilities for the encoder. SegFormer\cite{segformer} then proposed a lightweight multi-scale fusion MLP architecture that uses the transformer's backbone to achieve state-of-the-art performance on various datasets.

As shown in Figure \ref{compare}, transformer-based decoders have recently surpassed the convolution-based ones and achieved state-of-the-art performance, benefiting from their strong modeling capacity.
However, such approaches are computationally demanding, since the computational complexity of self-attention is quadratic to the number of patch tokens. 
In contrast, it is worth noting that the more light-weight MLP-based decoder has becoming a promising architecture, as its reasonable performance with few FLOPs. 
Despite its efficiency, we empirically find that it relies heavily on the receptive fields of the encoder, which is attributed to the standard channel-wise MLP operations can only fuse information along channel rather than spatial dimension, lacking the ability to perceive the contextual information.


To address this issue, we propose a cost-effective module, named Patch Rotate Module (PRM), which uses an parametric-free approach, performing rotate operations on pixels in the feature map to expand the receptive fields of the MLP architecture and fuse long-range contextual information. 
Specifically, we first use a Dynamic Channel Selection Module (DCSM) to dynamically divide the patch rotate partition and the reserved partition (i.e. the channels in the feature map that do not perform rotate operation). Then perform a regular rotate operation on the pixels of the feature map in the patch rotate partition. Finally, combine the feature map of the patch rotate partition with the reserved partition. In this way, a pixel in the same spatial location has the feature of different spatial locations under different channels. The final fully connected layer then has the ability to fuse long-range contextual information.

Based on the Dynamic Channel Selection Module (DCSM) and Patch Rotate Module (PRM), we design a light-weight MLP-based decoder, termed PRSeg. Meanwhile, in order to apply to different backbones, such as pyramid structured backbone (i.e., feature maps possessing hierarchical structure, e.g., ResNet-50\cite{ResNet}, SegFormer's encoder\cite{segformer}, Swin Transformer\cite{swin}, etc.) and straight backbone (i.e., the output feature maps are single-scale, e.g. ResNet-50-d8\cite{ResNet}, ViT\cite{ViT}, etc.), we designed two counterparts, called PRSeg-M (MultiScale) and PRSeg-S (SingleScale). The frameworks are shown in Figure \ref{PRSeg-M} and Figure \ref{PRSeg-S}, respectively.

Compared to the most related SegFormer \cite{segformer}, our PRSeg achieves obvious improvements under various backbones, especially for the light backbone variants (such as ResNet-50 \cite{ResNet}). For example, with FLOPs at just 30G, our PRSeg-M is +9.21$\%$ mIoU higher (42.36$\%$ vs. 33.15$\%$) than SegFormer on ADE20K dataset. When using ViT-Large \cite{ViT} as the backbone, PRSeg-S achieves 54.16$\%$ mIoU, outperforming previous state-of-the-art methods. For the Cityscapes dataset, when using ResNet-50 as the backbone, PRSeg-M achieved 80.84$\%$ mIoU at just 65 GFLOPs, which is +12.79$\%$ mIoU higher (80.84$\%$ vs. 68.05$\%$) than SegFormer. When using MiT-B5\cite{segformer} as the backbone, PRSeg-M achieves 84.42$\%$ mIoU.


\section{related work}
Since the milestone FCN \cite{FCN}, the encoder-decoder based architecture has been the cornerstone of semantic segmentation, where the encoder is used to extract features and the decoder aims to recover the dense prediction. 
Some works extract rich feature for semantic segmentation by designing powerful backbone, e.g. Swin Transformer\cite{swin}, CSWin Transformer\cite{CSWin}, etc. Some other works perform better fet-shot segmentation by designing backbone parameter fine-tuning, e.g., SVF\cite{SVF} by designing novel backbone small part parameter fine-tuning strategies to achieve better model generalization on learning new classes.
Many efforts have been focused on the design and development of decoders, which can be categorized into four groups, namely convolution-based decoder, attention-based decoder, transformer-based decoder and MLP-based decoder.
We introduce the development and characteristics of each type of segmentation decoder, and analyze their strengths and weaknesses in the following sub-sections.

\subsection{Convolution-based Decoder}

Although FCN \cite{FCN} leads the era of deep-learning based pixel-level prediction, its segmentation results are quite coarse.
Since then, lots of efforts have been made to improve the precision by enlarging the receptive fields \cite{pspnet, deeplab, Deeplabv2, deeplabv3, DeepLabv3+} and more comprehensive multi-scale fusion strategies \cite{semantic_fpn, upernet, SFNet, AlignSeg, FaPN}.
PSPNet \cite{pspnet} designed a pyramid spatial pooling module to aggregate multi-scale contextual information.
Sun et.al\cite{tcsvt2} proposed the Gaussian Dynamic Convolution (GDC) to fast and efficiently aggregate the contextual information and Meng et.al\cite{tcsvt3} proposed CNN-based multiple group cosegmentation network is first proposed to segment foregrounds employing two cues, the discriminative cue and the local-to-global cue. 
Ji et al.\cite{tcsvt5} proposed a CNN model based on the deformable convolutions to extract the non-rigid geometry aware features.
SFANet\cite{tcsvt4} proposed a Stage-aware Feature Alignment module (SFA) to align and aggregate two adjacent levels of feature maps effectively.
The Deeplab family \cite{Deeplabv2, deeplabv3, DeepLabv3+} developed the atrous spatial pyramid pooling (ASPP), containing multiple dilated convolutions with different dilated rates, to obtain larger receptive fields without the increase of computation. 
Inspired by the feature pyramid network \cite{FPN}, Kirillov et al. proposed Semantic FPN\cite{semantic_fpn} to gradually fuse multi-scale features in a top-down pathway for semantic segmentation task.
UperNet \cite{upernet} further enhanced the top-down feature fusion via pyramid spatial pooling for more contextual information.
To eliminate the misalignment issue during the aggregation between feature maps at different scales, recent SFNet \cite{SFNet}, AlignSeg \cite{AlignSeg} and FaPN \cite{FaPN} proposed to align spatial information by learning the transformation offsets according to the deviation of spatial position between different feature maps.

\subsection{Attention-based Decoder}
Benefited from the long-range modeling capacity, various attention modules have been developed for semantic segmentation  \cite{nonlocal,ann,danet,emanet,psanet,ocnet,CCNet,OCRNet,ISNet,SANet}.
The early non-local neural networks \cite{nonlocal} computed the response at a position as a weighted sum of the features at all positions.
ANN \cite{ann} proposed an asymmetric fusion Non-local block for fusing all features at one scale for each feature (position) on another scale.
DANet \cite{danet} via position attention module and channel attention module to adaptively integrate local features with their global dependencies.
CCNet \cite{CCNet} proposed a novel criss-cross attention module harvests the contextual information of all the pixels on its criss-cross path for each pixel.
OCNet \cite{ocnet} proposed an efficient interlaced sparse self-attention scheme to model the dense relations between any two of all pixels via the combination of two sparse relation matrices.
EMANet \cite{emanet} formulated the attention mechanism into an expectation-maximization manner and iteratively estimate a much more compact set of bases upon which the attention maps are computed.
PSANet \cite{psanet} used a self-adaptively learned attention mask for each position on the feature map connected to all the other ones. 
OCRNet \cite{OCRNet} proposed to improve the representation of each pixel by weighted aggregating the object region representations. 
ISNet \cite{ISNet} proposed to augment the pixel representations by aggregating the image-level and semantic-level contextual information, respectively. 
SANet \cite{SANet} proposed squeeze-and-attention modules impose pixel-group attention on conventional convolution by introducing an 'attention' convolutional channel, thus efficiently taking into account spatial-channel inter-dependencies.
CTNet\cite{ctnet} models the spatial and channel contextual relationships through the Spatial Contextual Module (SCM) and Channel Contextual Module (CCM) respectively. SSA\cite{SSA} has designed a novel semantic structure modeling module (SSM) to enable the generation of high quality CAMs during model inference.
ORDNet\cite{ordnet} efficiently captures short-, medium-, and long-range dependencies through the novel MiddleRange (MR) branch and Reweighed Long-Range (RLR) branch.



\subsection{Transformer-based Decoder}
The success of ViT \cite{ViT} has stimulated the community to introduce the transformer architecture into the design of the decoder for downstream tasks.
Feature Pyramid Transformer \cite{FPT} developed a top-down architecture with lateral connections to build high-level semantic feature maps at all scales.
Fully Transformer Networks\cite{FTN} extended the Semantic FPN \cite{semantic_fpn} to a transformer-based version for global dependencies.
EAPT \cite{EAPT} utilized deformable attention to learn an offset for each position in patches to obtain non-fixed attention information and cover various visual elements.
FSFormer \cite{FSFormer} proposed to perform non-progressive feature fusion across all the scales, and adaptively selected partial tokens as the keys for each scale.
Trans2Seg \cite{Trans2Seg} and Segmenter \cite{Segmenter} used transformer blocks to interact between class prototypes and feature maps.
MaskFormer \cite{maskformer} unified the semantic segmentation and instance segmentation tasks via a mask classification scheme.
Mask2Former \cite{Mask2Former} further strengthened the MaskFormer via masked attention, deformable attention and some training tricks.



\subsection{MLP-based Decoder}

Inspired by the surprising performance of pure MLP backbone networks \cite{mlpmixer, cyclemlp}, Xie et al. proposed SegFormer \cite{segformer} with only a few fully connected (fc) layers as the decoder for semantic segmentation.
\cite{tcsvt1} proposed Multi-head Mixer to explore rich context information from various subspaces.
Compared to the previous segmentation decoders, such MLP-based decoder is much more lightweight and efficient, which led to a new research hotspot.
As shown in Figure \ref{compare}, in comparison to DeepLabV3, SegFormer has 1.17 times the performance and only 0.3 times the FLOPs.
Compared to Segmenter, SegFormer achieves comparable performance to Segmenter with only 0.21 Params and 0.24 FLOPs.
However, the fc layer only enables channel-wise aggregation and lacks the ability of inter-pixel interaction. 
That is to say, its receptive fields are limited to each pixel.
Therefore, the existing MLP-based decoder \cite{segformer} relies heavily on the sufficient receptive fields of the encoder.
For example, when equipped with a light backbone, the performance of SegFormer dropped rapidly and lost its competitiveness.

Although the MLP-based decoder still has some issues to be solved, we believe it is a promising structure for efficient semantic segmentation and deserves further exploration.
In this paper, we are committed to exploring a novel semantic segmentation framework for MLP and hope that it will serve as an alternative for future semantic segmentation methods.


\section{Method}

In this section, we first introduce the framework of our PRSeg in Sec. \ref{sec:framework}. 
Then, we provide the motivation and detailed implementation of two key components, \textit{i.e.,} Dynamic Channel Selection Module (DCSM) and Patch Rotate Module (PRM) in Sec. \ref{sec:dcsm} and Sec. \ref{sec:PRM}, respectively. 
Finally, the loss function is presented in Sec. \ref{sec:loss}.

\begin{figure*}[tp]
\centering
\includegraphics[width=1.0\linewidth]{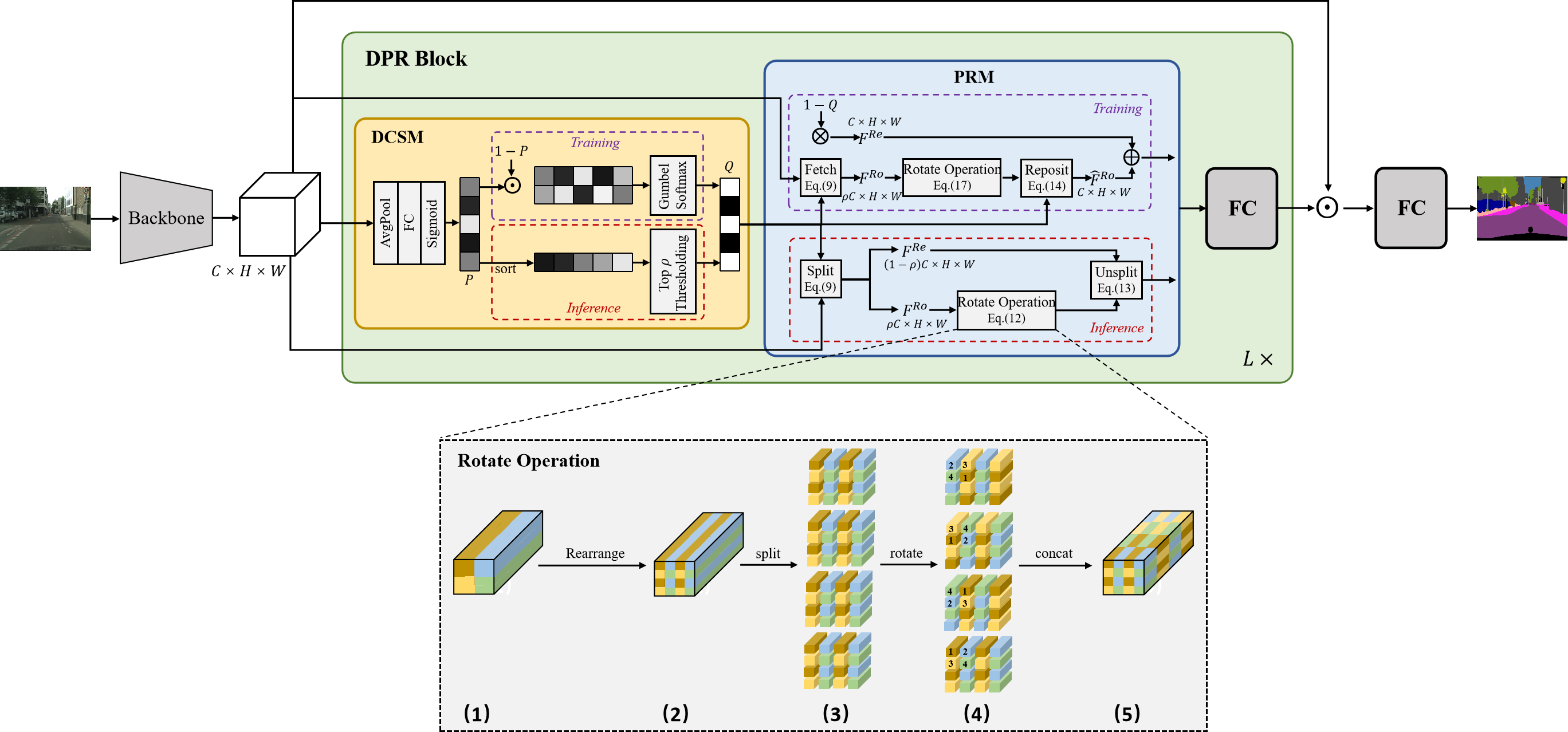} 
\caption{The overall architecture of our single-scale Patch Rotate MLP decoder (PRSeg-S). Its core design is the DPR-Block, which consists of a Dynamic Channel Selection Module (DCSM), a Patch Rotate Module (PRM) and a channel-wise Fully Connected (FC) layer.
}
\label{PRSeg-S} 
\end{figure*}

\subsection{Framework}
\label{sec:framework}
\begin{figure}[t]
\centering
\includegraphics[width=1.0\linewidth]{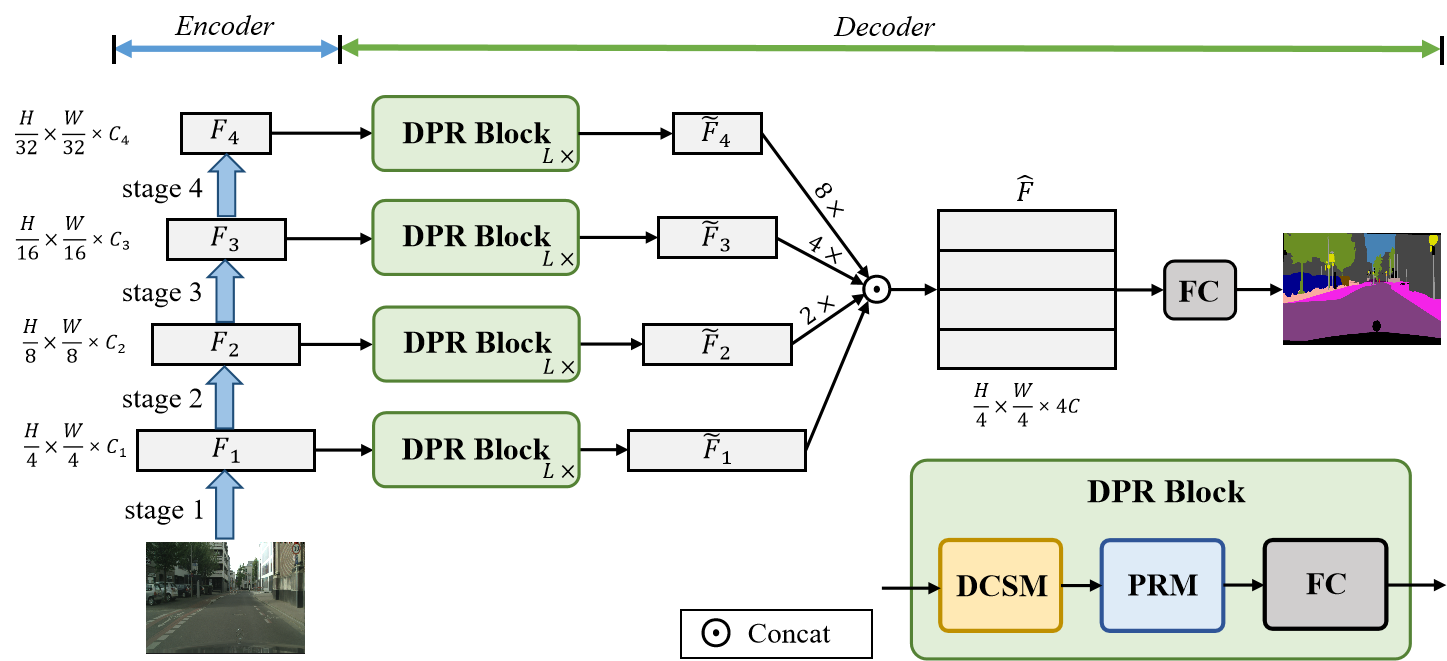} 
\caption{The overall architecture of our multi-scale Patch Rotate MLP decoder (PRSeg-M), which is a multi-scale extension of PRSeg-S with parallel branches , each of which is composed of several DPR-Blocks.
}
\label{PRSeg-M} 
\end{figure}

The pipeline of our Patch Rotate MLP Decoder (PRSeg) follows an encoder-decoder scheme, where the encoder is normally implemented via an off-the-shelf backbone (such as ResNet \cite{ResNet} and ViT \cite{ViT}) and the decoder is based on several proposed DPR-Blocks. 
The core of our DPR-Block is the Patch Rotate Module (PRM). 
It aims to extend the limited receptive fields of the channel-wise Fully Connected (FC) layer by performing the patch rotate operation on partial channels, which are selected by the Dynamic Channel Selection Module (DCSM), where the patch rotate operation spatially exchanges the information among neighbor pixels under a certain pattern.
Considering the popularity of both single-scale and multi-scale backbone networks\footnote{A single-scale backbone can be characterized by all the feature maps throughout the network having the same resolution. By contrast, the multi-scale backbones generate the hierarchical features, usually with 1/4, 1/8, 1/16, and 1/32 resolution of the input size.}, 
we design two different variants of PRSeg for better compatibility with them, termed PRSeg-S and PRSeg-M, which are illustrated in Figure \ref{PRSeg-S} and Figure \ref{PRSeg-M}, respectively.

Taking the PRSeg-S as an example, the input image $\bm{I} \in \mathbb{R}^{3 \times h \times w}$ is first passed through a single-scale backbone network (such as ViT \cite{ViT}) to extract the feature $\bm{F} \in \mathbb{R}^{C \times H \times W}$, where $H$, $W$ and $C$ denote the height, width and channel number, respectively.
Then, we stack $L$ DPR-Blocks to decode the feature. The output of the last DPR-Block $\bm{F}^{(L)}$ is concatenated with encoder output feature $\bm{F}$ to predict the pixel-level probability $Y \in [0,1]^{K \times H \times W}$ belonging to $K$ semantic class via a linear projection layer. The final segmentation result can be obtained by simply performing the argmax operation on $Y$ along the category dimension.
The forward pass of $l$-th DPR-Block can be formulated as follows:
\begin{align}
\label{eq:block-1}
& \bm{Q} = \text{DCSM} \big( \bm{F}^{(l-1)} \big), \\
& \bm{\widehat{F}}^{(l)} = \text{PRM} \big( \bm{F}^{(l-1)}, \bm{Q} \big), \\
& \bm{F}^{(l)} = \text{FC} \big( \bm{\widehat{F}}^{(l)} \big), 
\end{align}
where the input feature of the first DPR-Block $\bm{F}^{(0)} \in \mathbb{R}^{D \times H \times W}$ is implemented by $ \phi (\bm{F})$. $\phi: \mathbb{R}^{C} \mapsto \mathbb{R}^{D}$ is a linear projection layer to change the channel number.
Note that $\bm{F}^{(l-1)}$ and $\bm{F}^{(l)}$ have the same size.
The DCSM in Eq. \eqref{eq:block-1} is first used to predict a binary vector $\bm{Q} \in \{0,1\}^{C}$ as the indicator of whether performing the rotate operation for each channel according to the input content. $\bm{Q}_i=1, i \in \{1,...,C\}$ means the feature corresponding to $i$-th channel would be rotated in PRM, otherwise be reserved.
The PRM then applies the rotate operation on the selected channels (according to $\bm{Q}$) of the feature map $\bm{F}^{(l-1)}$.
The resulting transformed feature $\bm{\widehat{F}}^{(l)}$ is finally sent to FC layer for channel-wise projection.

Similarly, as shown in Figure \ref{PRSeg-M}, PRSeg-M is only a multi-scale extension of PRSeg-S. 
Specifically, given an input image $\bm{I} \in \mathbb{R}^{3 \times h \times w}$, it is first passed through a multi-scale backbone network (such as MiT \cite{segformer}) to extract the hierarchical features $\{ \bm{F}_i \}_{i=1}^S$, where $ \bm{F}_i \in \mathbb{R}^{C_i \times H_i \times W_i} $ represents the feature at $i$-th scale, and $H_i/W_i/C_i$ denote the height/width/channel number of $\bm{F}_i$. The total number of scales $S$ is usually set to 4.
Then, the feature at each scale is sent to several DPR-Blocks individually, and the outputs of all the parallel branches are integrated into a single feature via simple upsampling and concatenation. The following process to generate the final prediction is the same as PRSeg-S.

\subsection{Dynamic Channel Selection Module}
\label{sec:dcsm}

The Dynamic Channel Selection Module (DCSM) aims to dynamically select a $\rho$ percent subset of channels according to the individual content of each input image, for performing rotate operation.
Such purpose can also be achieved by simply selecting the first $\rho$ percent channels under a fixed pattern.
Considering the information required to be exchanged between pixels is distinct for different input images, we suppose our DCSM is a more adaptive manner, in which the channel subset is dynamically predicted according to the input contents.




Specifically, given a feature map $\bm{F} \in \mathbb{R}^{C \times H \times W}$, a spatial average pooling is first applied to obtain the global representation for each channel:
\begin{equation} 
\begin{aligned}
\bm{G}=\text{AvgPool}({\mathcal{F}}) \in \mathbb{R}^{C}.
\end{aligned}
\end{equation}
Then, we use a channel-wise fully connected layer to interact across channels following a sigmoid activation to constrain the value within $[0,1]$ for the probability meaning.
\begin{equation} 
    \begin{aligned}
\bm{P}=\text{Sigmoid}(\text{FC}({\bm{G}})) \in [0,1]^{C},
    \end{aligned}
\end{equation}
where the $i$-th element of $\bm{P}$ represents the preference degree of $i$-th channel for rotate operation. 
Finally, the probabilistic $\bm{P}$ is required to be converted to a binary indicator $\bm{Q} \in \{0,1\}^{C}$, where $\bm{Q}_i=1$ means $i$-th channel would be rotated, otherwise reserved.
Noth that such process differs for training and inference since the gradient back-propagation should be ensured during training.

\noindent
\textbf{Inference.} 
For simplicity, we directly perform Top-$\rho$ thresholding on $\bm{P}$. In detail, $\bm{P}$ is first sorted in a descending order, resulting $\bm{P}^{sort}$. Then, we take $\theta = \bm{P}^{sort}_{\rho}$ as the threshold to binarize the $\bm{P}$ as follows,
\begin{equation} 
\begin{aligned}
\bm{Q}_i = 
\begin{cases}
1,  & \bm{P}_i > \theta   \\
0,  & \bm{P}_i \leq \theta
\end{cases}
\end{aligned}
\end{equation}

\noindent
\textbf{Training.} 
The purpose of DCSM is to filter the channels used for rotation. However, if we employ softmax to obtain the probabilities of each channel, we need to determine the coordinates of the top 50$\%$ channels, and the coordinate sorting and sampling operations involved in this process are not differentiable, making it impossible to update the parameters of DCSM during backpropagation. 
To ensure the differentiability, we utilize the Gumbel-Softmax \cite{jang2016categorical} to sample a binary vector $\bm{Q}$ from the probability distribution $\bm{P}$ as follows,
\begin{align}
\label{eq:dcsm-1}
& \widehat{\bm{P}} = \text{Concat}(\bm{P}, 1-\bm{P}) \in [0,1]^{2\times C}, \\
\label{eq:dcsm-2}
& \bm{Q} = \text{Gumbel-Softmax}(\widehat{\bm{P}}).
\end{align}
Where, in Eq. \eqref{eq:dcsm-1}, $\bm{P}$ is first concatenated with its inverse to construct a 2-dimension probability distribution for each channel.
Then, the Gumbel-Softmax is applied on $\widehat{\bm{P}}$ for binarization according to the probability (Eq. \eqref{eq:dcsm-2}).

\subsection{Patch Rotate Module}
\label{sec:PRM}
The Patch Rotate Module (PRM) can be regarded as a pre-processing module of the fully connected (FC) layer to introduce the spatial-wise interaction among pixels. 
Different from the commonly-used spatial fusion operation (such as convolution and attention), we propose a more cost-effective manner to achieve such purpose, that is, performing rotate operation on the feature map corresponding to the channel subset selected by DCSM.


\noindent
\textbf{Inference.} 
We first divide the input feature map $\bm{F} \in \mathbb{R}^{C \times H \times W}$ into the reserved set $\mathcal{F}^{Re} = \{\bm{F}^{Re}_i\}_{i=1}^{N^{Re}}$ and rotated set $\mathcal{F}^{Ro} = \{\bm{F}^{Ro}_i\}_{i=1}^{N^{Ro}}$ along the channel dimension according to the binary indicator $\bm{Q}$ predicted by DCSM,
\begin{equation} 
\begin{aligned}
\label{eq:PRM_split}
\bm{F}_i \in 
\begin{cases}
\mathcal{F}^{Re},  & \bm{Q}_i = 0  \\
\mathcal{F}^{Ro},  & \bm{Q}_i = 1
\end{cases}
\end{aligned}
\end{equation}
where $i \in \{0,1,...,C\}$ is the channel index, $N^{Re} = (1 - \rho)C$ and $N^{Ro} = \rho C$. 
We denote $\sigma$ and $\tau$ as the one-to-one index mapping functions that satisfy $\bm{F}^{Re}_i = \bm{F}_{\sigma(i)}$ and $\bm{F}^{Ro}_i = \bm{F}_{\tau(i)}$.
Then, we concatenate all the spatial entries along channel dimension for $\mathcal{F}^{Re}$ and $\mathcal{F}^{Ro}$, respectively, resulting the reserved feature $\bm{F}^{Re} \in \mathbb{R}^{(1-\rho)C \times H \times W}$ and rotated candidate feature $\bm{F}^{Ro} \in \mathbb{R}^{\rho C \times H \times W}$,
\begin{align}
    \label{eq:concat-shift}
   & \bm{F}^{Re} = \text{Concat}(\bm{F}^{Re}_0, \bm{F}^{Re}_1, ..., \bm{F}^{Re}_{N^{Re}}), \\
   & \bm{F}^{Ro} = \text{Concat}(\bm{F}^{Ro}_0, \bm{F}^{Ro}_1, ..., \bm{F}^{Ro}_{N^{Ro}}).
\end{align}
Next, the rotate operation $f_{\text{rotate}}$ (refer to Eq. \eqref{eq:shift_op}) is performed on $\bm{F}^{Ro}$ to introduce information interaction between different pixels,
\begin{equation} 
    \begin{aligned}
    \label{eq:PRM_shift}
        \widetilde{\bm{F}}^{Ro} = f_{\text{rotate}}(\bm{F}^{Ro}).
    \end{aligned}
\end{equation}
Finally, we reorganized the reserved feature $\bm{F}^{Re}$ and rotated feature $\widetilde{\bm{F}}^{Ro}$ according to their corresponding channel indexes during division in Eq. \eqref{eq:PRM_split}.
\begin{equation} 
\begin{aligned}
\widehat{\bm{F}}_i =
\begin{cases}
\bm{F}^{Re}_{\sigma^{-1}(i)},  & \bm{Q}_i = 0  \\
\widetilde{\bm{F}}^{Ro}_{\tau^{-1}(i)},  & \bm{Q}_i = 1
\end{cases}
\end{aligned}
\end{equation}
Note that the output feature $\widehat{\bm{F}}$ of PRM has the same size of its input feature $\bm{F}$.

\noindent
\textbf{Training.} 
In order to allow the gradient back-propagation for all the entries in $\bm{Q}$, the training procedure of PRM also requires different implementations apart from the inference phase.
Specifically, given the input feature map $\bm{F} \in \mathbb{R}^{C \times H \times W}$, we first select the feature slices $\bm{F}_i$ of $\bm{F}$ that satisfy $\bm{Q}_i = 1$ as the rotated set $\mathcal{F}^{Ro} = \{\bm{F}^{Ro}_i\}_{i=1}^{N^{Ro}}$, \textit{i.e.,} if $\bm{Q}_i = 1$, $\bm{F}_i \in \mathcal{F}^{Ro}$.
$N^{Ro} = \rho C$, where $\rho C$ represents the total number of channels used for the rotated operation, $\rho$ is a probability value, which has the same meaning as the top $\rho$, and $i \in \{0,1,...,C\}$ is the channel index. We use $\tau$ to represent the one-to-one index mapping function that satisfy $\bm{F}^{Ro}_i = \bm{F}_{\tau(i)}$.
Then, as Eq. \eqref{eq:concat-shift}, the spatial entries of $\mathcal{F}^{Ro}$ are concatenated along channel dimension to obtain the rotated candidate feature $\bm{F}^{Ro} \in \mathbb{R}^{\rho C \times H \times W}$, which is passed through the rotate operation (same as Eq. \eqref{eq:PRM_shift}), resulting $\widetilde{\bm{F}}^{Ro}$.
Next, we reorganized the feature $\widetilde{\bm{F}}^{Ro} \in \mathbb{R}^{\rho C \times H \times W}$ to $\widehat{\bm{F}}^{Ro} \in \mathbb{R}^{C \times H \times W}$ by repositioning each slice back to its channel position in $\bm{F}$ and padding the other channels with zero value.
\begin{equation} 
\begin{aligned}
\widehat{\bm{F}}_i^{Ro} =
\begin{cases}
0,  & \bm{Q}_i = 0  \\
\widetilde{\bm{F}}^{Ro}_{\tau^{-1}(i)},  & \bm{Q}_i = 1
\end{cases}
\end{aligned}
\end{equation}
Meanwhile, the reserved feature ${\bm{F}}^{Re} \in \mathbb{R}^{C \times H \times W}$ can be simply obtained from $\bm{F}$ with $1-\bm{Q}$ as the filter,
\begin{equation} 
\begin{aligned}
\widehat{\bm{F}}^{Re} = \Theta(1-\bm{Q}_i) \otimes \bm{F},  
\end{aligned}
\end{equation}
where $\Theta:\mathbb{R}^{C} \mapsto \mathbb{R}^{C \times H \times W}$ is the repeat operation along spatial dimension. $\otimes$ denotes the element-wise multiplication.
Finally, the output transformed feature of PRM is obtained by element-wise summation between and $\widehat{\bm{F}}^{Re}$ and $\widehat{\bm{F}}^{Ro}$.
\begin{equation} 
\begin{aligned}
\widehat{\bm{F}} = \widehat{\bm{F}}^{Re} + \widehat{\bm{F}}^{Ro}.
\end{aligned}
\end{equation}

\noindent
\textbf{Rotate Operation.} 
\label{sec:shifto}
For efficient implementation, we divide the $\mathcal{F}^{Ro}$ into several different groups by space and channel. For a given $\mathcal{F}^{Ro}$ with a shape of $(K \times H \times W)$, k indicates the number of channels. Here we use a parameter $G_s$ to control the group size to be divided. $\mathcal{F}^{Ro}$ is first divided into $(G_s * G_s, K // (G_s * G_s), G_s, H // G_s, G_s, W // G_s)$.  then we rearrange each group of pixels on the space to $(G_s * G_s, K // (G_s * G_s), H // G_s, W // G_s, G_s * G_s)$, here the transformed tensor is denoted as $\mathcal{F}^{Ro'}$, a visual representation of the feature map at this point is shown in Figure \ref{PRSeg-S}. (2). 

Then we generate a rotated coordinate $\mathcal{I}$. The shape of this coordinate is equivalent to the $\mathcal{F}^{Ro'}$. The coordinates of a single pixel group $(G_s*G_s)$ are the same in the current channel group.  The coordinates of different channel groups are different. The coordinates of each pixel group are reordered clockwise between the different channel groups. So we do the patch rotate as in Figure \ref{PRSeg-S}. ($3 \rightarrow 4$). This process can be described as follows:

\begin{equation} 
    \begin{aligned}
    \label{eq:shift_op}
 \widehat{\mathcal{F}}^{Ro'}=\Phi(\mathcal{F}^{Ro'}, \mathcal{I}),
    \end{aligned}
\end{equation}
where $\Phi$ means patch rotate of $\mathcal{F}^{Ro'}$ according to the coordinate $\mathcal{I}$.

\subsection{Loss Function}
\label{sec:loss}
For fair comparisons, we adopt the commonly-used cross entropy as the loss function on the final segmentation result. 
In addition, we also employ a regularization loss to constrain the number of channels selected by our DCSM approaching the pre-defined proportion $\rho \in [0,1]$,
\begin{equation} 
\begin{aligned}
\mathcal{L}_{reg} = \frac{1}{L} \sum_{i=1}^{L} \Vert \rho -  \frac{1}{C} \sum_{j=1}^{C} \bm{Q}_{i}\Vert^2,
\end{aligned}
\label{loss1}
\end{equation}
where $L$ denotes the number of DPR-Block and $C$ is the total channel number of the feature map.
Above all, the total loss is the weighted combination of the cross-entropy loss and the regularization loss,
\begin{equation} 
\begin{aligned}
\mathcal{L} = \mathcal{L}_{ce} + \alpha \mathcal{L}_{reg} 
\end{aligned}
\label{loss2}
\end{equation}
where $\alpha$ is a hyper-parameter and set to 0.4 in our experiments. For more experimental results, see Figure \ref{fig:ablation} (d).



\section{Experiments}

In this section, we first introduce the datasets and implementation details in Sec. \ref{sec:exp-1}. 
Then, we compare our method with the recent state-of-the-arts in Sec. \ref{sec:exp-2}. 
Finally, in Sec. \ref{sec:exp-3}, extensive ablation studies and visualizations analysis are conducted to analyze the effect of key designs of our approach.


\subsection{Experimental Setup}
\label{sec:exp-1}

\noindent
\textbf{Datasets.} 
We evaluate our approach on the following datasets:

\begin{itemize}[leftmargin=0.3cm]
\item \textbf{ADE20K}\cite{ADE20K} is a very challenging benchmark with complex scenarios and high quality annotations including 150 categories and diverse scenes with 1,038 image-level labels, which is split into 20000 and 2000 images for training and validation.
\item \textbf{Cityscapes}\cite{Cityscapes} carefully annotates 19 object categories of high resolution urban landscape images. It contains 5K finely annotated images, split into 2975 and 500 for training and validation.
\item \textbf{COCO-Stuff 10K}\cite{COCOSTUFF}is a significant benchmark for scene parsing, consisting of 9000 training images and 1000 testing images, covering 171 categories.

\end{itemize}

\noindent
\textbf{Backbone.}
Our PRSeg is compatible with any backbone model. In this work, we use the well-known convolution-based ResNet-50 \cite{ResNet}, ResNet-101 and recently proposed transformer-based ViT-L \cite{ViT} and MiT-B5 \cite{segformer} to verify our good compatibility.
In addition, following the popular settings of ResNet in semantic segmentation community \cite{deeplab,Deeplabv2,deeplabv3,DeepLabv3+,pspnet,OCRNet,FCN}, we replace the first $7\times7$ with 3 consecutive $3\times3$ convolutions and use dilation convolutions at the last two stages to keep the output stride\footnote{The output stride denotes the ratio of the input image
spatial resolution to the final output resolution} of 8.

\noindent
\textbf{Protocols.} 
All the experiments are conducted on the 8 NVIDIA Tesla V100 GPUs (32 GB memory per-card) with PyTorch implement and mmsegmentation \cite{mmsegmentation} codebase.
For a fair comparison, we follow the training settings in previous works \cite{segformer,OCRNet,maskformer}, which are listed in Table \ref{tb:train_settings} for clarity. 
We simply apply the cross-entropy loss, and synchronized BN \cite{syncbn} to synchronize the mean and standard-deviation of BN \cite{bn} across multiple GPUs.
During the valuation, we report the widely-used mean intersection of union (mIoU) via both single-scale and multi-scale inference to measure the quality of segmentation results.
For the multi-scale inference, we apply the horizontal flip and average the predictions at multiple scales [0.5, 0.75, 1.0, 1.25, 1.5, 1.75].

\begin{table}[h]
\setlength\tabcolsep{1pt}
\centering
\caption{\label{tb:train_settings} Detailed training settings for each dataset with different backbone models. The ``Common'' item shows the settings shared for all the backbones.}
  \begin{tabular}{c|c|ccc}
    \hline
    \hline
    {}& {Training Settings}& {ADE20K}& {Cityscapes} & {COCO-Stuff 10k}\\
    \hline
                   &batch size         &16      &8  &16\\
                   &iterations         &160k    &80k &80k\\
    Common         &lr decay  & polynomial  & polynomial &polynomial\\
                   &random scale       &0.5$\sim$2  &0.5$\sim$2 &0.5$\sim$2\\
                   &random horizontal flip &0.5 &0.5 &0.5\\ 
                   
    \hline
    \multirow{6}{*}{ResNet}   
          &optimizer          &SGD    &SGD &SGD\\
                   &learning rate      &0.01    &0.01 &0.01\\
          &weight decay       &0.0005    &0.0005 &0.0005\\
                   &optimizer momentum &0.9    &0.9 &0.9\\
                   &warmup             &no    &no &no\\
                   &random crop         &$512\times512$    &$768\times768$ &$512\times512$\\
    \hline
                   &optimizer          &AdamW    &AdamW\\
                   &learning rate      &0.00006    &0.00006\\
                   &weight decay       &0.01    &0.01\\
    MiT-B5         &optimizer momentum &(0.9, 0.999)    &(0.9, 0.999)\\
                   &warmup schedule    &linear    &linear\\
                   &warmup iterations       &1500    &1500\\
                   &random crop         &$640\times640$    &$1024\times1024$\\
    \hline
                   &optimizer          &AdamW    &\\
                   &learning rate      &0.00002    &\\
                   &weight decay       &0.01    &\\
    ViT-L          &optimizer momentum &(0.9, 0.999)    &\\
                   &warmup schedule    &linear    &\\
                   &warmup iterations       &1500    &\\
                   &random crop         &$640\times640$    &\\
    \hline
    \hline
    \end{tabular}
\end{table}

\noindent
\textbf{Reproducibility.} 
Our method is implemented in PyTorch (version $\ge$ 1.5) and trained on 8 NVIDIA Tesla V100 GPUs with a 32 GB memory per-card. We used public codebase mmsegmentation \cite{mmsegmentation} for all our experiments.

\subsection{Comparisons with the state-of-the-art}
\label{sec:exp-2}

\noindent
\textbf{ADE20K.} 
Table \ref{sotaade} reports the comparison with the state-of-the-art methods on the ADE20K validation set. 
When ResNet-50 is used as the backbone, our PRSeg-M is +9.21$\%$ mIoU higher (42.36$\%$ vs. 33.15$\%$) than SegFormer\cite{segformer} with the same input size ($512 \times 512$) and our PRSeg-S achieves 44.40$\%$ mIoU. While recent methods \cite{segformer, Segmenter} showed that using a larger resolution $(640 \times 640)$ can bring more improvements. To make a fair comparison with SegFormer, we use SegFormer's backbone (i.e. MiT-B5), and train it under the same setting, our PRSeg-M is +1.17$\%$ mIoU higher (52.97$\%$ vs. 51.80$\%$) than SegFormer. And in order to compare with the latest methods, we added experiments using ViT \cite{ViT} as the backbone, our PRSeg-S is +0.56$\%$ mIoU higher (54.16$\%$ vs. 53.60$\%$) than Segmenter\cite{Segmenter}. It is worth mentioning that the FLOPs of our PRSeg are much lower than the state-of-the-arts.
Additionally, we conducted FPS tests for each method on the Nvidia 1050-Ti GPU, and the results are shown in Table \ref{fps}. It is worth noting that our PRSeg-M has only 1.14 FPS lower compared to SegFormer-ResNet-50, while its performance is 11.25$\%$ mIoU higher (44.40$\%$ vs. 33.15$\%$).

\begin{table}[h]
\setlength\tabcolsep{5pt}
\centering
\caption{Comparison with the state-of-the-art methods on the ADE20K dataset. ``SS'' and ``MS'' indicate single-scale inference and multi-scale inference, respectively.}
    \label{sotaade}
  \begin{tabular}{l|c|c|c|c|c}
    \hline
    \hline
    {Method} & {Backbone} & {GFLOPs} & {Params} & {SS} & {MS}\\
    \hline
    FCN \cite{FCN}             & ResNet-50  &198 &50M &36.10 & 38.08 \\
    EncNet \cite{EncNet}       & ResNet-50  &141 &36M &40.10 & 41.71 \\
    CCNet \cite{CCNet}         & ResNet-50  &201 &50M &42.08 & 43.13 \\
    ANN \cite{ann}             & ResNet-50  &185 &46M &41.74 & 42.62 \\
    PSPNet \cite{pspnet}       & ResNet-50  &179 &49M &42.48 & 43.44 \\
    OCRNet \cite{OCRNet}       & ResNet-50  &153 &37M &42.47 & 43.55 \\
    DeepLabV3 \cite{deeplabv3} & ResNet-50  &270 &68M &42.66 & 44.09 \\
    DMNet \cite{DMNet}         & ResNet-50  &196 &53M &43.15 & 44.17\\
    SegFormer\cite{segformer}  & ResNet-50  &30 &25M &32.96 & 33.15\\
    \rowcolor{gray}
    PRSeg-M(ours)             & ResNet-50  &30&30M &\textbf{41.58} & \textbf{42.36}\\
    \rowcolor{gray}
    PRSeg-S(ours)             & ResNet-50  &110&26M &\textbf{43.98} & \textbf{44.40}\\
    \hline
    SegFormer\cite{segformer}  & MiT-B5     &82 &82M &51.00 & 51.80\\
    \rowcolor{gray}
    PRSeg-M(ours)             & MiT-B5     &82 &82M &\textbf{52.05} & \textbf{52.97}\\
    \hline
    DPT\cite{DPT}              & ViT-L      &328 &338M &49.16 & 49.52\\
    UperNet\cite{upernet}      & ViT-L      &710 &354M &48.64 & 50.00\\
    SETR \cite{SETR}           & ViT-L      &332 &310M &50.45 & 52.06\\
    MCIBI \cite{MCIBI}         & ViT-L      &-   &-    &-     & 50.80\\
    Segmenter \cite{Segmenter} & ViT-L      &380 &342M &51.80 & 53.60\\
    \rowcolor{gray}
    PRSeg-S(ours)             & ViT-L      &329&309M&\textbf{53.21} & \textbf{54.16}\\
    \hline
    \hline
    \end{tabular}
\end{table}

\begin{figure*}[tp]
\centering
\includegraphics[width=1\linewidth]{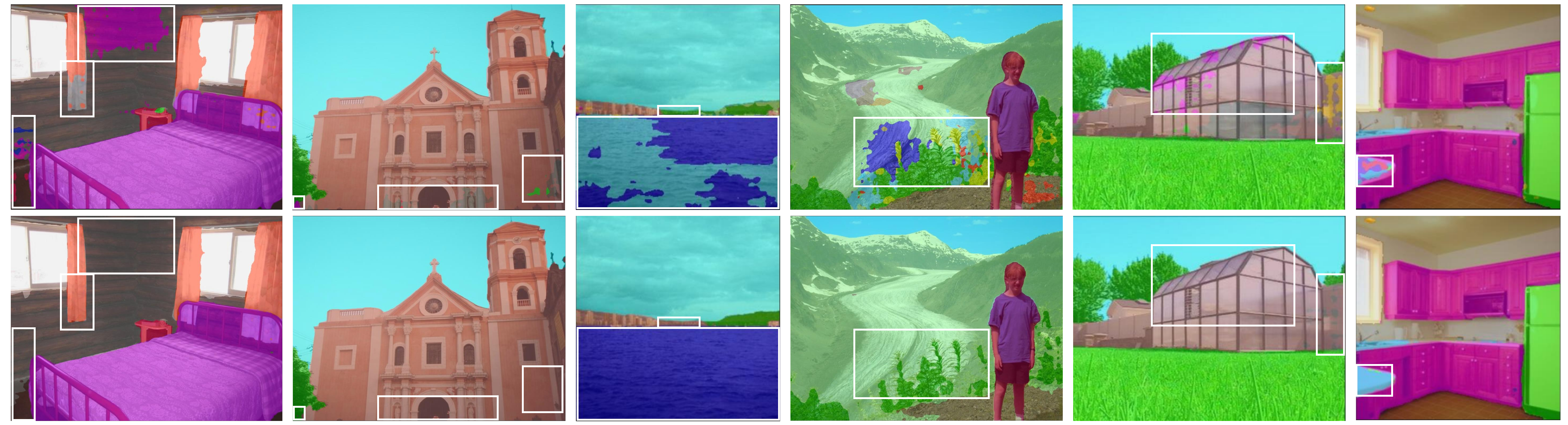} 
\caption{
Qualitative visualization of SegFormer (top) and our PRSeg-M (bottom).
The examples are chosen from ADE20K dataset.
Compared to SegFormer, our PRSeg-M reduces long-range context fusion errors as highlighted in white box. 
}
\label{visfig} 
\end{figure*}

\begin{table}[h]
\setlength\tabcolsep{5pt}
\centering
\caption{Comparison with the state-of-the-art methods on the Cityscapes validation set.}
    \label{sotacitys}
  \begin{tabular}{l|c|c|c|c|c}
    \hline
    \hline
    {Method} & {Backbone} & {GFLOPs} & {Params} & {SS} & {MS}\\
    \hline
    FCN \cite{FCN}             & ResNet-50  &396 &50M &73.61 & 74.24 \\
    EncNet \cite{EncNet}       & ResNet-50  &282 &36M &77.94 & 79.13 \\
    ANN \cite{ann}             & ResNet-50  &370 &46M &77.34 & 78.65 \\
    PSPNet \cite{pspnet}       & ResNet-50  &357 &49M &78.55 & 79.79 \\
    CCNet \cite{CCNet}         & ResNet-50  &401 &50M &79.03 & 80.16 \\
    DMNet \cite{DMNet}         & ResNet-50  &391 &53M &79.07 & 80.22\\
    DeepLabV3 \cite{deeplabv3} & ResNet-50  &540 &68M &79.32 & 80.57 \\
    SegFormer\cite{segformer}  & ResNet-50  &65 &25M &59.33 & 68.05\\
    \rowcolor{gray}
    PRSeg-S(ours)               & ResNet-50  &246 &26M &\textbf{79.54} & \textbf{80.81}\\
    \rowcolor{gray}
    PRSeg-M(ours)               & ResNet-50  &65 &30M &\textbf{79.16} & \textbf{80.84}\\
    \hline
    SegFormer\cite{segformer}  & MiT-B5     &208 &82M &82.40 & 84.00\\
    \rowcolor{gray}
    PRSeg-M(ours)               & MiT-B5  &208 &82M &\textbf{83.07} & \textbf{84.42}\\
    \hline
    \hline
    \end{tabular}
\end{table} 

\begin{table}[!h]
\setlength\tabcolsep{11.9pt}
\centering
\caption{Comparison with the state-of-the-art methods on the COCO-Stuff 10K dataset.}
  \begin{tabular}{l|c|c|c}
    \hline
    \hline
    \multirow{2}{*}{Method} &\multirow{2}{*}{Venue} & \multirow{2}{*} {Backbone}    & mIoU \\
    &  & & (MS) \\
    \hline
     PSPNet \cite{pspnet}         &CVPR17 &ResNet-101 & 38.86 \\
     SVCNet \cite{SVCNet}         &CVPR19 &ResNet-101 & 39.60 \\
     DANet \cite{danet}           &CVPR19 &ResNet-101 & 39.70 \\
     EMANet \cite{emanet}         &ICCV19 &ResNet-101 & 39.90 \\
     SpyGR \cite{SpyGR}           &CVPR20 &ResNet-101 & 39.90 \\
     ACNet \cite{ACNet}           &ICCV19 &ResNet-101 & 40.10 \\
     OCRNet \cite{OCRNet}         &ECCV20 &HRNet-W48 & 40.50 \\
     GINet \cite{GINet}           &ECCV20 &ResNet-101 & 40.60 \\
     RecoNet \cite{RecoNet}       &ECCV20 & ResNet-101 & 41.50\\
     ISNet \cite{ISNet}           &ICCV21 & ResNeSt-101& 42.08\\
    \rowcolor{gray}
    PRSeg-S(ours)           &- & ResNet-101& \textbf{41.65}\\
    \rowcolor{gray}
    PRSeg-M(ours)           &- & ResNet-101& \textbf{42.43}\\
    \hline
    \hline
    \end{tabular}
    \label{sotacoco}
\end{table}

\begin{table*}[t]
\setlength\tabcolsep{6pt}
\centering
\caption{Analysis of FPS. Except for SegFormer and PRSeg-M, which use straight architecture ResNet-50 (dilation=8) as their backbone, all other methods adopt the pyramidal architecture ResNet-50.}
    \label{fps}
  \begin{tabular}{l|c|c|c|c|c|c|c|c}
    \hline
    \hline
    Method & FCN\cite{FCN}  &PSPNet\cite{pspnet}  &OCRNet\cite{OCRNet}  &DeepLabV3\cite{deeplabv3}  &DMNet\cite{DMNet}  &SegFormer\cite{segformer} &PRSeg-S &PRSeg-M\\
    \hline
    FPS    &3.58 &3.79 &4.41 &3.00 &3.32 &9.23 &3.25 &8.09\\
    \hline
    \hline
    \end{tabular}
\end{table*}

\noindent
\textbf{Cityscapes.} 
Table \ref{sotacitys} shows the comparative results on the Cityscapes validation set. Due to the high efficiency of the MLP decoder, our PRSeg-M achieves 80.84$\%$ mIoU with 65 GFLOPs when using ResNet-50 as the backbone, with an average of only 16$\%$ of the computation and only 50$\%$ number of parameters compared to other methods. Our PRSeg-M is +12.69$\%$ mIoU higher (80.84$\%$ vs. 68.15$\%$) than SegFormer\cite{segformer} with the same input size ($768 \times 768$).  When using the more powerful MiT-B5\cite{segformer} as a backbone, we achieved 84.42$\%$ mIoU, which is +0.42$\%$ mIoU higher (84.42$\%$ vs. 84.00$\%$) than SegFormer, and outperforms the state-of-the-art methods.


\noindent
\textbf{COCO-Stuff 10K.} 
Table \ref{sotacoco} shows the comparison of PRSeg with SOTA method on the COCO dataset. Where PRSeg-M achieves 42.43$\%$ mIoU, which is +0.35$\%$ mIoU higher than the previous SOTA method ISNet\cite{ISNet}.

\noindent
\textbf{Summary.} By comparing the experiments of SegFormer, we can find that the reason for the good performance of SegFormer using MLP as decoder is that the perceptive filed of transformer backbone is large enough, so the decoder part does not need a large perceptive filed to fuse the long-range context information. However, when using a backbone with an average perceptive filed like ResNet-50, SegFormer's MLP decoder performance drops dramatically. PRSeg's Patch Rotate Module is designed to address this issue.

Furthermore, in Table \ref{sotacitys}, when SegFormer's MLP decoder uses ResNet-50 as the backbone, the single-scale and multi-scale results are 59.33$\%$ mIoU and 68.05$\%$ mIoU, respectively. This enhancement is a bit unusual.

Table \ref{segformerscale} presents the results of our multi-scale validated ablation studies on SegFormer. Our experiments demonstrate that only the backbone can fuse context information, as the perceptive field of SegFormer-head is limited to a single pixel. In contrast, the perceptive field of ResNet-50 is local and fixed. Therefore, when the input image resolution is reduced, the perceptive field of ResNet-50 relatively increases, leading to an improved performance of SegFormer-head. Conversely, as the input image resolution increases, the perceptive field of ResNet-50 becomes relatively smaller, segformer-head does not have the ability to fuse long-range context information and resulting in a significant performance degradation. This experiment highlights the drawback of a plain MLP decoder, which is its inability to effectively fuse long-range contextual information. It underscores the necessity of designing an MLP decoder that can integrate surrounding information to improve its performance.

\begin{table}[h]
\setlength\tabcolsep{5pt}
\centering
\caption{SegFormer-ResNet-50 multi-scale inference performance on ADE20K dataset.}
    \label{segformerscale}
  \begin{tabular}{l|c|c|c|c|c|c}
    \hline
    \hline
    Scale Ratio & 0.5  &0.75  &1.00  &1.25  &1.50  &1.75\\
    \hline
    {mIoU}        &71.31 &71.34 &63.40 &61.87 &60.21 &58.26\\
    \hline
    \hline
    \end{tabular}
\end{table} 

\subsection{Ablation Study}
\label{sec:exp-3}

In this subsection, we conduct ablation studies under PRSeg-S with ResNet-50 as backbone on the ADE20K dataset. 

\noindent\textbf{Effect of Each Component in DPR-Block.} 
To investigate the performance enhancements yielded by each component in our DPR-Block, we conducted experiments on various combinations of DCSM, PRM, and FC layers.
As illustrated in Table \ref{tb:ablation_component}, the utilization of only two fully connected layers resulted in a mIoU of 35.12$\%$. Upon adding the Patch Rotate Module (PRM), performance was elevated by 8.35$\%$ mIoU (43.47$\%$ vs. 35.12$\%$), with significant improvements attributed to PRM without any increase in FLOPs and Params.
Further inclusion of the Dynamic Channel Selection Module (DCSM) led to a modest performance gain of 0.51$\%$ mIoU, culminating in an overall mIoU of 43.98$\%$.
This ablation studies effectively demonstrates the efficacy of the proposed method and thoroughly highlights the substantial performance improvements achieved by expanding the MLP decoder receptive field.

\begin{table}[t]
\caption{Ablation study on the effect of each component in our DPR-Block.}
\centering
\begin{tabular}{ccc|c}
\hline
\hline
DCSM        & PRM         & FC layer       & mIoU (SS)   \\ 
\hline
            &             & \checkmark     &  35.12\\
            & \checkmark  & \checkmark     &  43.47\\
\checkmark  & \checkmark  & \checkmark     &  43.98\\
\hline
\hline
\end{tabular}
\label{tb:ablation_component}
\end{table}

\begin{figure*}[t] 
\centering
    \begin{minipage}[c]{0.24\textwidth}
    \centering
    \includegraphics[width=\textwidth]{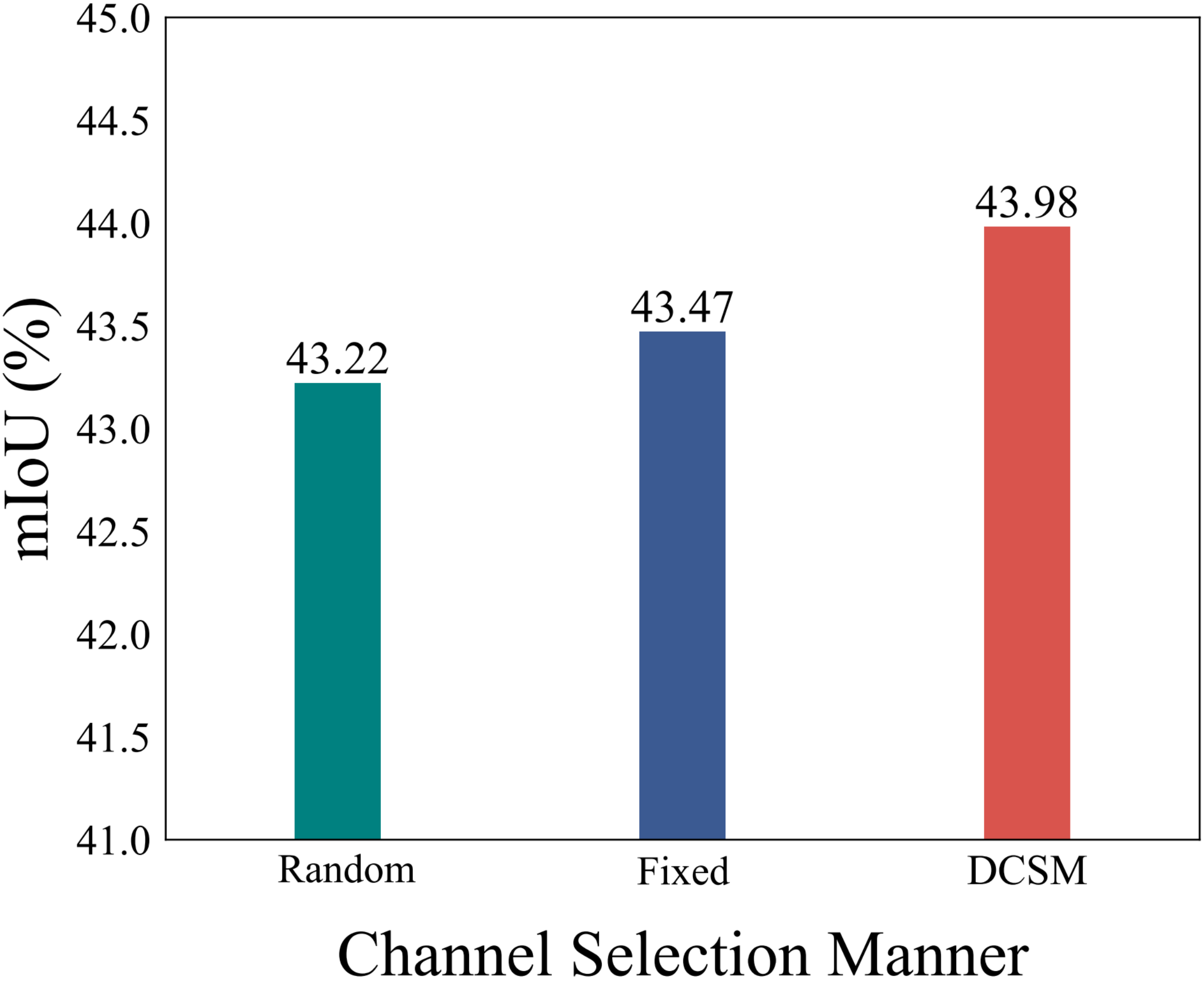}
    \subcaption*{\enspace \thinspace (a)}
    \label{fig:ablation_dcsm_vs_other}
    \end{minipage}
    \hspace{0.001\textwidth}
    \begin{minipage}[c]{0.24\textwidth}
    \centering
    \includegraphics[width=\textwidth]{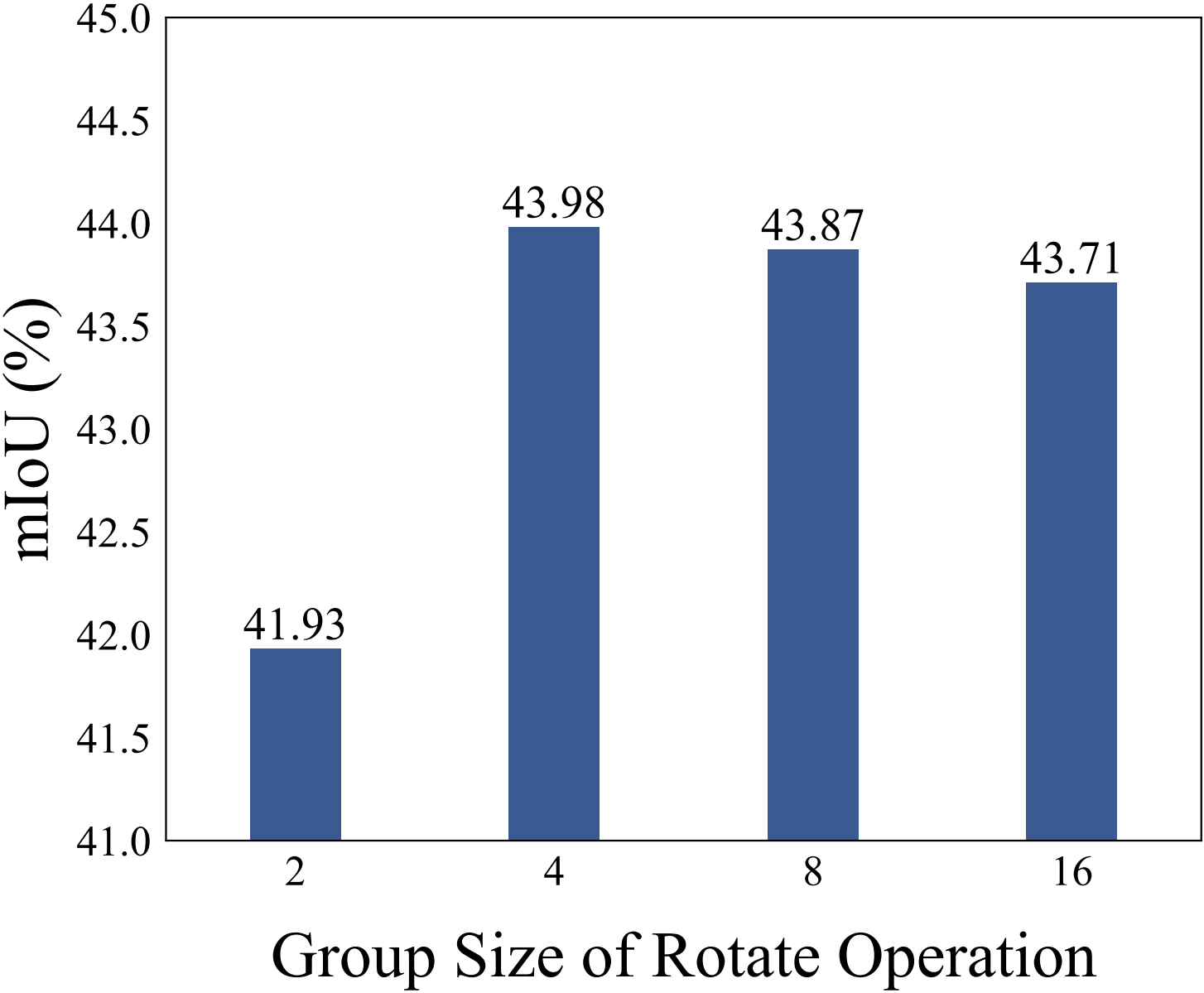}
    \subcaption*{\enspace \thinspace (b)}
    \label{fig:ablation_shift_group_size}
	\end{minipage}
    \hspace{0.001\textwidth}
    \begin{minipage}[c]{0.238\textwidth}
    \centering
    \includegraphics[width=\textwidth]{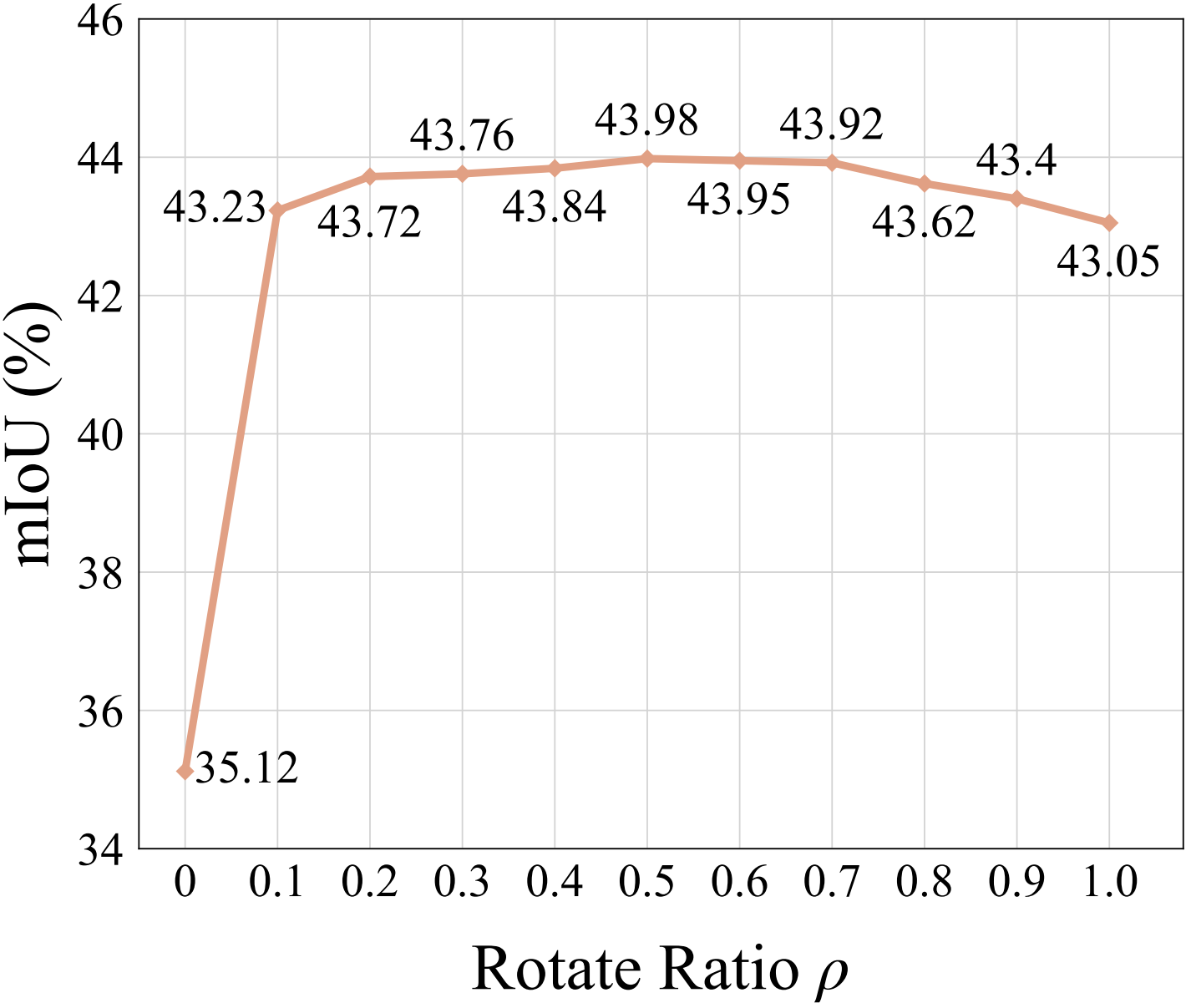}
    \subcaption*{\quad \thinspace (c)}
    \label{fig:ablation_shift_ratio}
    \end{minipage}
    \hspace{0.001\textwidth}
    \begin{minipage}[c]{0.24\textwidth}
    \centering
    \includegraphics[width=\textwidth]{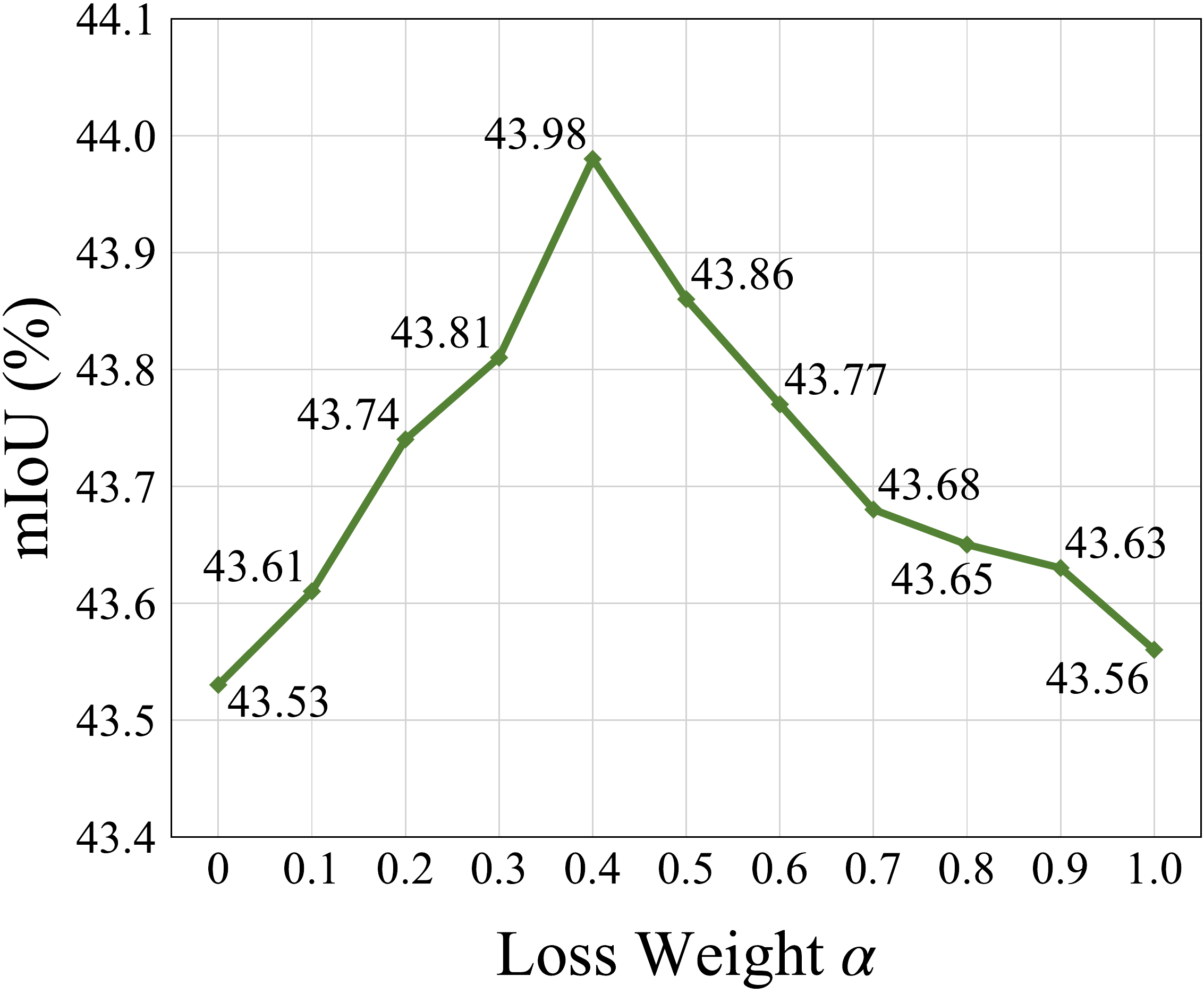}
    \subcaption*{\quad \thinspace (d)}
    \label{fig:ablation_loss_weight}
    \end{minipage}
    \caption{\label{fig:ablation}
	(a) Comparisons with different channel selection manners. 
	(b) Effect of group size in rotate operation.
	(c) Effect of the ratio of rotated channels.
	(d) Effect of the weight of regulation loss in total loss function.
    }
\end{figure*}

\noindent\textbf{DCSM vs. Others.}
As described in Sec. \ref{sec:dcsm}, our DCSM adaptively selects a 50\% subset of channels according to the image content.
Here, we compare our proposed DCSM with two baseline channel selections to verify its effectiveness: 
(i) \textit{``Random''} means randomly selecting 50\% channels; and 
(ii) \textit{``Fixed''} means directly selecting the former 50\% channels.
As shown in Figure \ref{fig:ablation} (a), our DCSM achieves the best 43.98\% mIoU, which is +0.76\% mIoU and +0.51\% mIoU higher than the random and fixed manner, respectively. 
Unlike the static selection strategies of ``Fixed'' and ``Random'', DCSM is a dynamic modeling approach. With DCSM, the network can adaptively filter channels for patch rotation operations. DCSM can be adaptively adjusted for different samples to achieve better performance gains.


\noindent\textbf{Patch Rotation Strategy.}
Table \ref{prs} presents a comparison of patch rotation strategies on ADE20K. It is evident that our rotation strategy outperforms the random rotation strategy by 0.45$\%$ mIoU. Moreover, both rotation strategies lead to a significant improvement in performance for the single-pixel receptive field of the MLP decoder (without a patch rotation strategy has only 35.12$\%$ mIoU).

\begin{table}[!h]
\centering
\caption{Ablation study on patch rotation strategy in our Rotate Operation.}
  \begin{tabular}{c|c}
    \hline
    \hline
    {Patch Rotation Strategy}& {mIoU(SS)}\\
    \hline
    Random     &43.53\\
    Ours     &43.98\\
    \hline
    \hline
    \end{tabular}
    \label{prs}
\end{table}

\noindent\textbf{Rotate Ratio.} 
The rotate ratio $\rho$ represents the proportion of channels used for the patch rotate operation. 
A larger rotate ratio can bring more information interaction among neighbor pixels, while may also lead to noise context aggregation. 
We study the effect of rotate ratio in Figure \ref{fig:ablation} (c).
It can be seen that the performance changes with rotate ratio in a unimodal pattern within a wide range of 9\%.
Specifically, it dramatically increases from 35.12\% over $\rho=0\sim0.5$, peaking at 43.98\%, and then falls back to 43.05\% at $\rho=1$.
The significant performance improvement observed for values of $\rho$ between 0 and 0.1 underscores the importance of enhancing the MLP decoder's perception field and provides compelling evidence for the efficacy of the proposed rotated module.
Note that the rotate ratio is not the larger the better, which may attribute to the information disruption caused by forcibly exchanging excessive channels between pixels.
The best mIoU is achieved at a rotate ratio of 0.5, thus we set the rotate ratio to 0.5 by default.


\noindent\textbf{Group Size of rotate operation.}
Figure \ref{fig:ablation} (b) shows the effect of the group size in the rotate operation, which indicates how many groups are divided on the space and channel. For more details, please refer to the rotate operation in Sec. \ref{sec:shifto}. The best performance of the model was achieved with $G_{s}$=4, so we chose it for all experiments.


\noindent \textbf{Number of DPR-Block.} 
The depth of PRSeg was extensively evaluated through ablation experiments listed in Table \ref{depth}, ranging from 1 to 6. It is noteworthy that the Patch Rotate Module does not involve any parameters, thus each additional block results in an increase in both computation and the number of network parameters by one fully connected layer. As the depth increases, the network's performance improves. Considering the trade-off between efficiency and performance, we opted for Block numbers=2 in all experiments.
\begin{table}[h]
\centering
\caption{Ablation study on the number of blocks. A block consists of a Dynamic Channel Selection Module (DCSM), a Patch Rotate Module (PRM) and a channel-wise Fully Connected (FC) layer. The following results are obtained on ADE20K with a decoder of dimension 512.}
  \begin{tabular}{c|c|c|c}
    \hline
    \hline
    {Block numbers}& {GFLOPs}& {Params}& {mIoU}\\
    \hline
    1              &108.9    &25.7M    &42.90\\
    \textbf{2}              &\textbf{110.0}    &\textbf{26.3M}    &\textbf{43.98}\\
    3              &111.0    &26.8M    &43.88\\
    4              &112.1    &27.3M    &43.79\\
    5              &113.2    &27.8M    &43.85\\
    6              &114.3    &28.3M    &44.51\\
    \hline
    \hline
    \end{tabular}
    \label{depth}
\end{table}

\noindent \textbf{Dimension of Decoder.} 
The ablation study of decoder dimension is shown in Table \ref{dimension}. We can see from the experimental data that as dimension increases (i.e., from 192 to 2048), the performance of the network, the number of parameters, and the amount of computation increase. With the trade-off of efficiency and performance, we chose C=512 for all experiments.
\begin{table}[h]
\centering
\caption{Accuracy as a function of the MLP dimension C in the decoder on ADE20K. We use ResNet-50 as the backbone and measure the parameters and GFLOPs at crop size = $512 \times 512$.}
  \begin{tabular}{c|c|c|c|c|c}
  \hline
  \hline
  \multirow{2}{*}{C}& 
  \multicolumn{2}{c|}{GFLOPs}& 
  \multicolumn{2}{c|}{Params}&
  \multirow{2}{*}{mIoU}\\
  \cline{2-5} 
         & Encoder & Decoder & Encoder & Decoder & \\
  \hline
    192  &         &2.3      &         &0.6M    &42.09\\
    384  &         &5.9      &         &1.7M    &42.84\\
    \textbf{512}  &         &\textbf{9.0}      &         &\textbf{2.7M}    &\textbf{43.98}\\
    768  &101.0    &16.6     &23.5M    &5.2M    &44.01\\
    1024 &         &26.4     &         &8.5M    &44.12\\
    1536 &         &52.5     &         &17.5M   &44.33\\
    2048 &         &87.2     &         &29.7M   &44.56\\
    \hline
    \hline
    \end{tabular}
    \label{dimension}
\end{table}

\begin{figure}[!h]
\centering
\includegraphics[width=1\linewidth]{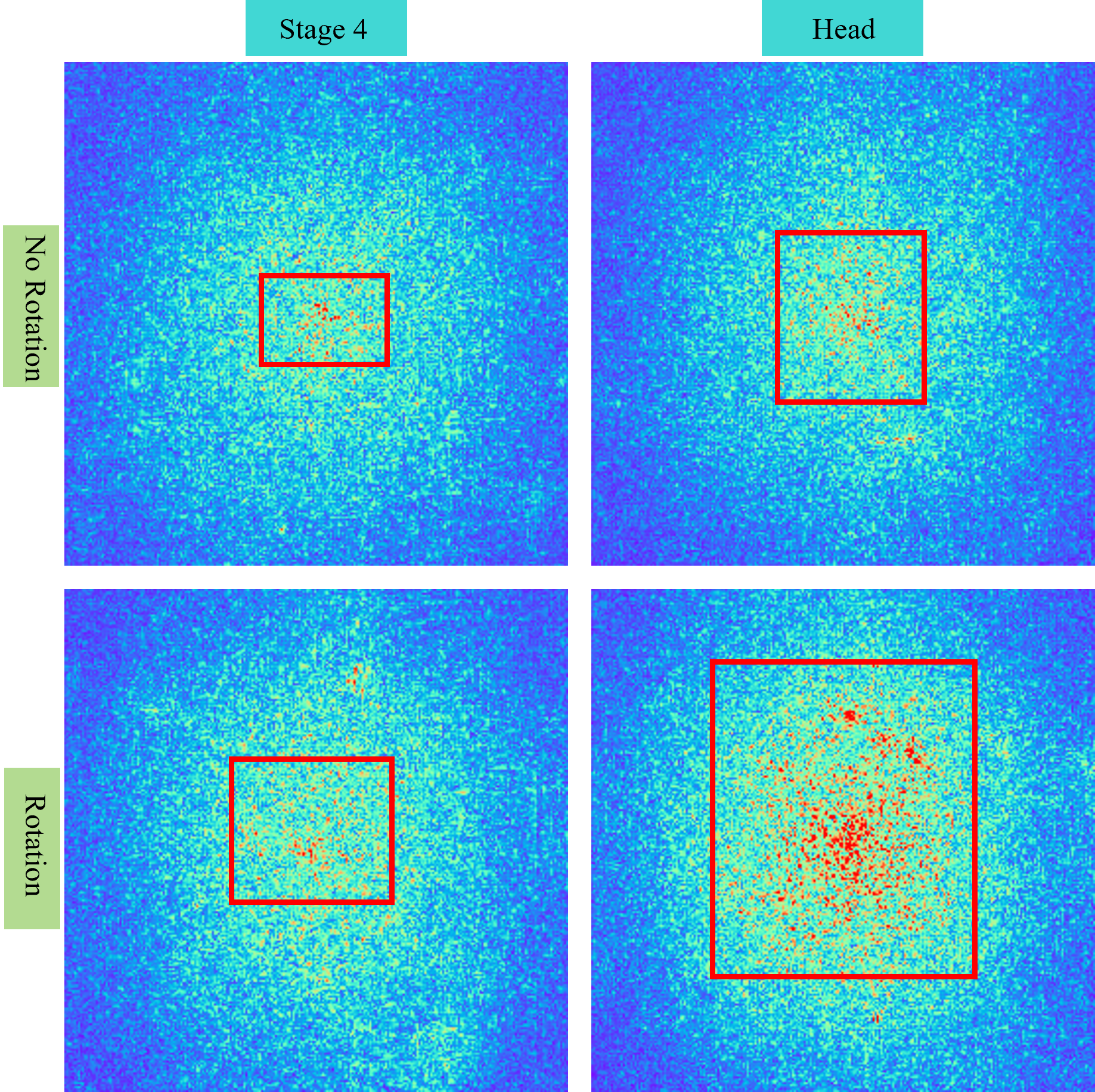} 
\caption{Effective Receptive Field (ERF) on ADE20K (average over 100 images). ERFs of the stage4 and head are visualized. Best viewed with zoom in.}
\label{erf}
\end{figure}

\noindent \textbf{Loss Weight $\alpha$.} 
To study the effect of Loss Weight $\alpha$, we test different weights $\alpha =\{0.0, 0.2, 0.4, 0.6, 0.8, 1.0\}$. 
As shown in Figure \ref{fig:ablation} (d), it can be seen that $\alpha = 0. 4$ can yield the best accuracy (43.98$\%$ mIoU).

\noindent \textbf{Effect of Final Concatenation.} 
The final concatenation is the concatenation of the backbone output with the last DPR-Block output on the channel.
The final concatenation avoids any particular channel responses to be over-amplified or suppressed \cite{FaPN}, which is increased by 0.32$\%$ mIoU to 43.98$\%$ mIoU.

\noindent \textbf{Effective Receptive Field.}  
For dense prediction tasks (e.g., segmentation and detection), fusing more context information with a larger receptive field has been a central issue. We use the effective receptive field (ERF)\cite{erf} as a visualization tool to explain the effectiveness of the proposed rotate  operation. As shown in Fig. \ref{erf}, we use PRSeg-S-ResNet-50 as the base model, where ``No Rotation'' represents the absence of DCSM and Rotate Operation in the base model, and ''Rotation'' represents the standard PRSeg-S-ResNet-50. The results indicate a significant improvement in the perceptive field of the head following rotation, providing evidence for the efficacy of the proposed method.

\section{Conclusion}
In this paper, we propose a Patch Rotate MLP decoder (PRSeg), a simple and efficient pure MLP decoder for semantic segmentation.  
It consists multiple Dynamic Patch Rotate Blocks (DPR-Blocks),
in each DPR-Block, which consists of a Dynamic Channel Selection Module, a Patch Rotate Module and a Fully Connected layer. It overcomes the previous problem that the perceptive field of MLP as a decoder is only one pixel. In the above experiments, it can be found that MLP networks also have powerful modeling capabilities. As an option for future semantic segmentation model development, PRSeg could be used as a baseline for the MLP decoder. Moreover, we hope to inspire the rethinking of MLP networks in the field of computer vision.

\section*{Acknowledgments}
The study is partially supported by National Natural Science Foundation of China under Grant (U2003208) and Key R \& D Project of Xinjiang Uygur Autonomous Region(2021B01002).


{\small
\bibliographystyle{ieee_fullname}
\bibliography{egbib}
}

\vfill

\end{document}